\documentclass{article}

\usepackage{microtype}
\usepackage{graphicx}
\usepackage{booktabs} 
\usepackage{float}
\usepackage{flafter}
\usepackage{hyperref}
\usepackage{multirow}

\usepackage[accepted]{icml2024}

\usepackage{algpseudocode}

\usepackage{enumitem}

\usepackage{url}

\usepackage[export]{adjustbox}
\usepackage[caption=false,font=scriptsize]{subfig}

\usepackage{amsmath}
\usepackage{amssymb}
\usepackage{mathtools}
\usepackage{amsthm}
\usepackage{fixmath}
\usepackage{amsmath}
\usepackage{amsfonts}
\usepackage{amssymb}
\usepackage{mathtools}

\DeclareMathOperator*{\argmin}{arg\,min}

\icmltitlerunning{Memory-Efficient Vision Transformers: An Activation-Aware Mixed-Rank Compression Strategy}

\begin{document}

\twocolumn[
\icmltitle{Memory-Efficient Vision Transformers: An Activation-Aware Mixed-Rank Compression Strategy}




\begin{icmlauthorlist}
\icmlauthor{Seyedarmin Azizi}{usc}
\icmlauthor{Mahdi Nazemi}{usc}
\icmlauthor{Massoud Pedram}{usc}
\end{icmlauthorlist}

\icmlaffiliation{usc}{Department of Electrical and Computer Engineering, University of Southern California, Los Angeles, United States}

\icmlcorrespondingauthor{Seyedarmin Azizi}{seyedarm@usc.edu}

\icmlkeywords{Machine Learning, ICML}

\vskip 0.3in
]



 \printAffiliationsAndNotice{} 

\begin{abstract}
As Vision Transformers (ViTs) increasingly set new benchmarks in computer vision, their practical deployment on inference engines is often hindered by their significant memory bandwidth and (on-chip) memory footprint requirements. This paper addresses this memory limitation by introducing an activation-aware model compression methodology that uses selective low-rank weight tensor approximations of different layers to reduce the parameter count of ViTs. 
The key idea is to decompose the weight tensors into a sum of two parameter-efficient tensors while minimizing the error between the product of the input activations with the original weight tensor and the product of the input activations with the approximate tensor sum. This approximation is further refined by adopting an efficient layer-wise error compensation technique that uses the gradient of the layer's output loss. 
The combination of these techniques achieves excellent results while it avoids being trapped in a shallow local minimum early in the optimization process and strikes a good balance between the model compression and output accuracy. 
Notably, the presented method significantly reduces the parameter count of DeiT-B by 60\% with less than 1\% accuracy drop on the ImageNet dataset, overcoming the usual accuracy degradation seen in low-rank approximations.
In addition to this, the presented compression technique can compress large DeiT/ViT models to have about the same model size as smaller DeiT/ViT variants while yielding up to 1.8\% accuracy gain.
These results highlight the efficacy of our approach, presenting a viable solution for embedding ViTs in memory-constrained environments without compromising their performance. 

\end{abstract}

\section{Introduction}
\label{submission}
Vision Transformers (ViTs), highlighted in key studies like \cite{DBLP:conf/iclr/DosovitskiyB0WZ21}, are recognized for their strong performance in various computer vision tasks. These tasks include image classification \cite{DBLP:conf/iclr/DosovitskiyB0WZ21, DBLP:conf/icml/TouvronCDMSJ21, DBLP:conf/iccv/LiuL00W0LG21}, object detection \cite{DBLP:conf/eccv/CarionMSUKZ20, DBLP:conf/nips/FangLWFQWNL21}, semantic segmentation \cite{DBLP:journals/corr/abs-2102-04306, DBLP:conf/iccv/StrudelPLS21}, and multi-modal virtual assistance \cite{anonymous2024jailbreak, liu2023llava, shayegani2023survey}. However, the broader implementation of these transformers is notably hampered by their extensive parameter requirements, resulting in significant memory footprints.

In response to this challenge, model compression has emerged as the quintessential strategy for facilitating the deployment of models characterized by a high parameter count. Predominant techniques in this domain encompass model pruning \cite{DBLP:conf/iclr/ZhuG18, zhu2021vision, DBLP:conf/iclr/LiuSZHD19, DBLP:conf/aaai/0004HWCCC22, DBLP:conf/nips/ChenCGYZW21}, token pruning \cite{DBLP:conf/eccv/KongDMMNSSYRTQW22}, quantization \cite{DBLP:conf/nips/LiuWHZMG21, DBLP:conf/eccv/YuanXCWS22}, and knowledge distillation \cite{DBLP:conf/icml/TouvronCDMSJ21, DBLP:conf/eccv/WuZPLXFY22}. These methodologies collectively aim to reduce the computational and storage burden, thereby enabling the efficient deployment of these advanced neural network architectures in resource-constrained environments.

Within the array of model compression techniques, low-rank approximation stands out as a particularly effective strategy for model compression, due to two reasons: (1) It has a solid theoretical foundation with proven optimality, as shown in \cite{eckart1936approximation}, and (2) Its structured application directly translates to hardware efficiency and implementation ease. The effectiveness of this method is demonstrated in significant studies like \cite{DBLP:conf/bmvc/JaderbergVZ14, DBLP:conf/ijcnlp/NoachG20, DBLP:conf/iclr/HsuHCLSJ22}.

Despite its potential, the na\"ive application of low-rank decomposition to the weights of ViTs often leads to a significant decline in performance, specifically when targeting higher compression rates. This issue arises primarily because the parameters of transformer-based models are typically not inherently suited to low-rank structures, as detailed in \cite{DBLP:conf/iclr/HsuHCLSJ22}. This underscores the need for a more nuanced and tailored approach in the application of low-rank approximation techniques to ViTs.


To rectify the challenges associated with the application of low-rank decomposition to ViT weights, this paper introduces an innovative method for the decomposition of a pre-trained weight matrix. Our approach involves the decomposition of the weight matrix into a sum of two low-rank matrices, each contributing distinctively to the accurate reconstruction of the original matrix. Given a pre-trained weight matrix \(\mathbold{W}\), we aim to approximate it as:
\begin{equation}
    \mathbold{W} \approx \mathbold{\mathbold{U}\mathbold{V}^T} +\mathbold{Z}
\end{equation}
where \(\mathbold{\mathbold{U}}\) and \(\mathbold{V}\) are low-rank matrices, and \(\mathbold{Z}\) is also a matrix with a constrained number of parameters. Importantly, \(\mathbold{Z}\) must be designed for efficient hardware implementation and effective in approximating \(\mathbold{W} - \mathbold{U}\mathbold{V}^T\). This configuration is strategically designed to maintain the aggregate parameter count of approximation matrices significantly lower than that of the original, pre-trained weight matrix.

The proposed method, which will be detailed later, ensures that the product \(\mathbold{\mathbold{U}\mathbold{V}^T}\) and \(\mathbold{Z}\) provide distinct yet complementary contributions to the reconstruction of \(\mathbold{W}\). Specifically, \(\mathbold{\mathbold{U}\mathbold{V}^T}\) captures the \textbf{principal energy} of the matrix through singular value decomposition (SVD), while \(\mathbold{Z}\) aims to offset the \textbf{residual error} from SVD using a layer-wise gradient-based optimization process.
%

%
%

Our approach for low-rank decomposition of weight matrices of ViTs is supported by two important observations: (1) statistics of the input feature map for each layer play a key role in influencing the approximation error associated with the parameters of the layer (as also discussed in \cite{DBLP:conf/aaai/YuW23}), and (2) layers of ViTs exhibit different susceptibilities to low-rank approximation, that is, aggressive rank reduction in some layers results in notable performance degradation at the model output. These insights form the foundation of our strategy, described as \textbf{activation-aware mixed-rank compression}, allowing for a smoother and tailored reduction in the number of parameters, thereby preserving the principal energy of the original ViT model's weight matrices. The contributions of this paper may be summarized as follows:
\begin{itemize}[left=-1pt]
\item We formulate model compression as a general optimization problem that aims to find a low-rank approximation for each weight matrix in the ViT while minimizing the aggregate energy loss across all weight matrices.
\item Our investigation delves into the impact of activation awareness in the application of singular value decomposition (SVD). We present a practical and highly effective methodology that incorporates input activations for the approximation of weight matrices, enhancing the approximation quality and capturing the principal energy contents of each layer.
\item In terms of SVD implementation, we employ a strategic, gradual rank reduction approach, which judiciously assigns varying ranks to different layers within the model. The method is based on the fitness of the layers to low-rank approximation, and picks a layer which if its approximate tensor undergoes a parameter count reduction, the amount of information loss is minimum.
\item To address the approximation error that is inherent in activation-aware SVD, we formulate a layer-specific gradient-based optimization problem. This approach aims to minimize the reconstruction error at each layer by decomposing the original weight matrix into a combination of the SVD result and a low parameter-count matrix, denoted as \(\mathbold{Z}\). This crucial step serves to recuperate the energy loss encountered as a result of compression.
\item We extend our methodology to various ViTs, conducting comprehensive experiments. These experiments yield compelling results in both accuracy and compression, demonstrating significant parameter count reduction without compromising model performance. Although our primary goal is reducing the memory footprint, we show in appendix \ref{flops} that adopting low-rank approximation not only does not hurt the computational efficiency but also improves it. Thus, our method introduces no memory-computation trade-off. 
\end{itemize}

\section{Background}

A ViT architecture comprises a collection of identical blocks, each block comprising four layers, including the attention Query-Key-Value (QKV) layer, Attention Projection (AttnProj) layer, and two feed-forward Up Projection and Down Projection layers realized as Multi-Layer Perceptrons (MLP1 and MLP2). 

Given a layer's pre-trained weight matrix \(\mathbold{W} \in \mathbb{R}^{n \times d}\), singular value decomposition (SVD) may be used to factorize it into  \(\mathbold{W} = \mathbold{U} \mathbold{\Sigma} \mathbold{V}^T\). Here, \(\mathbold{U} \in \mathbb{R}^{n \times n}\) and \(\mathbold{V} \in \mathbb{R}^{d \times d}\) represent unitary matrices comprising the left and right singular vectors, respectively, while \(\mathbold{\Sigma} \in \mathbb{R}^{n \times d}\) is a diagonal matrix containing the singular values. To approximate \(\mathbold{W}\) with a rank \(r\) using SVD, the process involves retaining only the top \(r\) largest singular values and their corresponding singular vectors, resulting in an approximation \(\hat{\mathbold{W}} \approx\ \mathbold{U}_r \mathbold{\Sigma}_r {\mathbold{V}_r}^T\). As established in \cite{eckart1936approximation}, this specific rank-\(r\) approximation is optimal, yielding minimal error amongst all potential rank \(r\) matrices.

Aligned with previous studies on implementing low-rank structures in neural networks \cite{DBLP:conf/bmvc/JaderbergVZ14, DBLP:journals/corr/TaiXWE15, DBLP:conf/cvpr/YuLWT17}, the linear transformation of a layer may be approximated as follows: \begin{equation} \mathbold{O} = \mathbold{X}\mathbold{W} + \mathbold{b} \approx \mathbold{X}(\mathbold{U}_r \mathbold{\Sigma}_r {\mathbold{V}_r}^T) + \mathbold{b} \end{equation}
where $\mathbold{X}$ is the layer's input activation, $\mathbold{b}$ is the bias, and \(\mathbold{O}\) is the output.

Following the methodology in \cite{DBLP:conf/iclr/HsuHCLSJ22}, the singular values can be integrated into the left and right singular vectors. This integration results in a total post-approximation parameter count of \(n \times r + d \times r\). \textbf{Throughout the paper we use the \(n_l\) and \(d_l\) to refer to the dimensions of the original weight matrix of layer \(l\).}

The exploration of low-rank approximation, particularly within the realms of transformers and ViTs, has been an active area of research. For instance, \cite{DBLP:conf/iclr/HsuHCLSJ22} effectively utilized Fisher Information as a means to evaluate the significance of the weight matrices in transformers, subsequently refining the objective of SVD to incorporate gradient awareness, thereby boosting its efficacy in model compression. Further advancements in this field are evident in the work of \cite{DBLP:conf/aaai/YuW23}, where eigendecomposition is applied to the covariance of the layer's output, revealing its enhanced suitability for low-rank approximation.
Lastly, \cite{kumar2022vision} introduced a hybrid methodology that combines pruning techniques for feed-forward layers with low-rank decomposition for attention blocks, showcasing an innovative approach to model compression.

Despite these advancements, a common challenge persists across these methods: a significant drop in accuracy when high levels of compression are applied through low-rank structures, such as a 50\% reduction in parameter count. This accuracy decline is largely attributed to the inability of these approaches to adequately compensate for the perturbations and energy loss induced by low-rank approximation. This highlights the necessity for more refined techniques that can effectively balance the trade-off between model compression and performance retention.

Another related work \cite{DBLP:conf/icml/LiYZLHCZ23} approximates the pre-trained weight matrix as a summation of a low-rank and sparse matrix, where the low-rank matrix captures the coherent part of the matrix, and a sparse matrix approximates the remaining incoherent residual; they demonstrated that this decoupling makes the matrix easy to prune. Despite this, since the SVD decomposition is directly applied to the weight matrix, they suffer from considerable energy losses (removals of large singular values), which results in noticeable performance degradation in high compression rates, highlighting the need for more sophisticated methods that can preserve the essential characteristics of the weight matrix.

\section{Problem Formulation and Solution Methodology}
\label{method}
In our setting, given a pre-trained neural network model \(\mathcal{M}\), which has \(L\) layers, each represented by a weight matrix \(\mathbold{W}_l\), we are looking for a configuration of ranks \(\mathbold{r} = [r_1, r_2,\cdots, r_L]\) such that when each layer \(l\) is approximated by rank \(r_l\), the summation of the layers' normalized errors across model $\mathcal{M}$ (denoted by Err($\mathcal{M}$)) is minimized:
\begin{equation}
\begin{aligned}
\label{mainproblem}
\mathrm{Err}(\mathcal{M}) =   \min_{\mathbold{r}} &\sum_{l}\min_{\hat{\mathbold{W}_l}}E_{\mathbold{X}_l} \left[\frac{\parallel \mathbold{X}_l\hat{\mathbold{W}_l} - \mathbold{X}_l\mathbold{W}_l \parallel_F^2 }{\parallel \mathbold{X}_l\mathbold{W}_l \parallel_F^2 }\right]\\
\textrm{s.t.} \quad  &\rho(\hat{\mathbold{W}_l}) \leq r_l \quad \forall l,   \\
  & \psi(\mathcal{M},\mathbold{r}) \geq \alpha \\
\end{aligned}
\end{equation}

Here, \(\mathbold{X}_l\) denotes the input activation for a layer \(l\), \(\mathbold{W}_l\) denotes the layer's pre-trained weight matrix, \(\hat{\mathbold{W}_l}\) is the low-rank approximation of the weight matrix, constrained from above by rank \(r_l\), \(\rho()\) denotes a function that returns the rank of the input matrix, \(E_{\mathbold{X}_l}()\) denotes the expectation value over the set of all \(\mathbold{X}_l\) activations, and the function \(\psi(\mathcal{M}, \mathbold{r})\) computes the ratio of parameter count reduction of the model \(\mathcal{M}\) after approximating each original weight matrix \(\mathbold{W}_l\) by a low-rank matrix \(\hat{\mathbold{W}_l}\) of rank \(r_l\).
The main differentiating point between this formulation and prior work that deal with layer-wise output reconstruction (e.g., \cite{DBLP:conf/icml/HubaraNHBS21, DBLP:conf/nips/FrantarA22, DBLP:conf/icml/NagelABLB20}) is the incorporation of the sum of total normalized reconstruction errors across all layers, which effectively accounts for the interaction of layers in terms of compressibility.


If we define the quantity inside inner minimization as \(\varepsilon_{l} = \min_{\hat{\mathbold{W}_l}}E_{\mathbold{X}_l} \left[\frac{\big(\parallel \mathbold{X}_l\hat{\mathbold{W}_l} - \mathbold{X}_l\mathbold{W}_l \parallel_F^2 \big)}{\big(\parallel \mathbold{X}_l\mathbold{W}_l \parallel_F^2 \big)}\right]\), then \(\varepsilon_{l}\) captures at layer \(l\) the Frobenius norm of the output difference between the compressed representation and its uncompressed counterpart \textbf{in a normalized sense}. The Frobenius norm \(\| \mathbold{A} \|_F^2\) is defined as \(\sum_{i,j}a_{ij}^2\) and is equivalent to the sum of the squares of the singular values \(\sigma_i\) of \(A\) \cite{DBLP:books/cu/HJ2012}, representing the \textbf{energy of the matrix}.

This optimization seeks a rank configuration vector \(\mathbold{r} \) assigning ranks to layers, which minimizes the summation of normalized energy losses across all layers. The Min-Sum-Min nature of this optimization problem aims to achieve a balanced, minimal energy loss landscape across all layers, thus preserving the model's inherent characteristics. This is a challenging optimization problem because the objective function is non-convex. Moreover, since the rank of each layer can be any integer number, the total number of possible configurations \(\mathbold{r} \) that achieve the desired compression ratio \(\alpha\) is exponentially large. Essentially, any configuration meeting the following condition is a feasible candidate:
\begin{equation}
    \psi(\mathcal{M},\mathbold{r} ) = \frac{\sum_{l}n_l \times d_l}{\sum_{l}(n_l\times r_l +d_l\times r_l)} \geq \alpha.
\end{equation}

This problem, in its general form, is NP-hard, indicating a high level of complexity and computational challenge. 

To efficiently solve this problem, we have developed a multi-step heuristic flow, as explained below.
Section \ref{act-aware lra} presents an \textbf{Activation-Aware Low-Rank Approximation} technique where, for a given rank configuration vector \(\mathbold{r} \) and layer \(l\), the inner minimization of \eqref{mainproblem} is solved. That is, we compute the minimum energy loss \(\varepsilon_{l}\) and the corresponding optimized low-rank weight matrices \(\mathbold{U}_l\) and \(\mathbold{V}_l\). This aspect of the framework leverages the characteristics of each layer's input activations to determine the most effective low-rank approximation.

Section \ref{MRMC} introduces a \textbf{Mixed-Rank Compression} technique. To solve the outer minimization in \eqref{mainproblem}, this strategy defines the rank configuration vector \(\mathbold{r} \) through a methodical, greedy local neighborhood search. It continuously integrates feedback from the activation-aware SVD process, ensuring that the rank distribution across layers is optimally aligned with their remaining compression potential and performance constraints.

Finally, section \ref{gbopt} proposes a \textbf{Layer-wise Error Compensation} technique, whose goal is to balance the residual errors introduced by the SVD. By employing a layer-wise gradient-based optimization technique, this technique finds a new matrix \(\mathbold{Z}_l\) to be added to the product of \(\mathbold{U}_l\) and \(\mathbold{V}_l\) for each layer \(l\), further refining the low-rank approximation and enhancing the overall quality of the approximation.

\subsection{Activation-Aware Low-Rank Approximation} \label{act-aware lra}
First and foremost, we reformulate the reconstruction error based on our proposed approach:
\begin{equation}
    \label{layerwise}
     \varepsilon_{l} = \min_{\mathbold{U}_l,\mathbold{V}_l,\mathbold{Z}_l}E_{\mathbold{X}_l} \left[\frac{\big(\parallel \mathbold{X}_l(\mathbold{U}_l\mathbold{V}_l^T+\mathbold{Z}_l) - \mathbold{X}_l\mathbold{W}_l \parallel_F^2 \big)}{\big(\parallel \mathbold{X}_l\mathbold{W}_l \parallel_F^2 \big)}\right]
\end{equation}

Given the challenges in directly computing the expectation, we follow a methodology similar to \cite{DBLP:conf/nips/FrantarA22} by constructing a proxy dataset. This dataset comprises samples of input activations for each layer, and we approximate the expectation with the average over these samples in the proxy dataset. A critical insight we've observed is the importance of considering the input activation during the application of singular value decomposition (SVD) to maximally preserve a layer's output, as targeted in  \eqref{layerwise}.

Re-examining Problem \ref{layerwise}, if we initially set \(\mathbold{Z}_l\) to zero and the value of \(r_l\) is predetermined, then for a single input image \(\mathbold{X}_l^i\), the problem can be reformulated as \(\min_{\mathbold{U}_l,\mathbold{V}_l}\| \mathbold{X}_l^i(\mathbold{U}_l\mathbold{V}_l^T) - \mathbold{X}_l^i\mathbold{W}_l \|_F^2\) (notice that the denominator is not a function of \(\mathbold{U}_l\) or \(\mathbold{V}_l\)). For this special case, we can find the optimal \(\mathbold{U}_l\) and \(\mathbold{V}_l\) that minimize the reconstruction error as follows:
\begin{equation}
  \label{equation4}
  \mathbold{X}_l^i\mathbold{U}_l\mathbold{V}_l^T = SVD_{r_l}(\mathbold{X}_l^i\mathbold{W}_l) = \mathbold{U}_l^*\mathbold{\Sigma}_l^*{\mathbold{V}_l^*}^T
 \end{equation}
In this context, \(SVD_k\) denotes the application of SVD while retaining the largest \(k\) singular values \(\mathbold{\Sigma}_l^*\) and their associated left and right singular vectors (\(\mathbold{U}_l^*, \mathbold{V}_l^*\)), respectively. A feasible solution for \(\mathbold{U}\) and \(\mathbold{V}\) could be as follows:
\begin{equation}
    \label{svd}
    \mathbold{U}_l = {\mathbold{X}_l^i}^\dag \mathbold{U}_l^*\sqrt{\mathbold{\Sigma}_l},  \: \: \: \: \: \: \: \mathbold{V}_l = \sqrt{\mathbold{\Sigma}_l}\mathbold{V}_l^*,  
\end{equation}
where \(\mathbold{X}_l^{i^\dag}\) denotes the pseudo-inverse of the input activation.

This formulation considers the influence of single input activation. When utilizing a proxy dataset comprising \(N\) samples extracted from the original dataset rather than a single image, it becomes necessary to generate a representative input \(\mathbold{X}_\mathrm{rep}\) for the application of \eqref{svd}. To simplify this process, one practical approach is to use the average of the samples in the proxy dataset. This representative sample \(\mathbold{X}_\mathrm{rep}\) can be calculated as \(\mathbold{X}_\mathrm{rep} = \frac{1}{N}\sum_{i=1}^{N}\mathbold{X}^i\). This method ensures that \(\mathbold{X}_\mathrm{rep}\) effectively captures the characteristics of the proxy dataset, providing a more holistic representation for the application of SVD. Then, we can use \(\mathbold{X}_\mathrm{rep}\) in \eqref{svd} to obtain \(\mathbold{U}\) and \(\mathbold{V}\).



\begin{figure}[tb]
    \centering
    \includegraphics[width=\columnwidth]{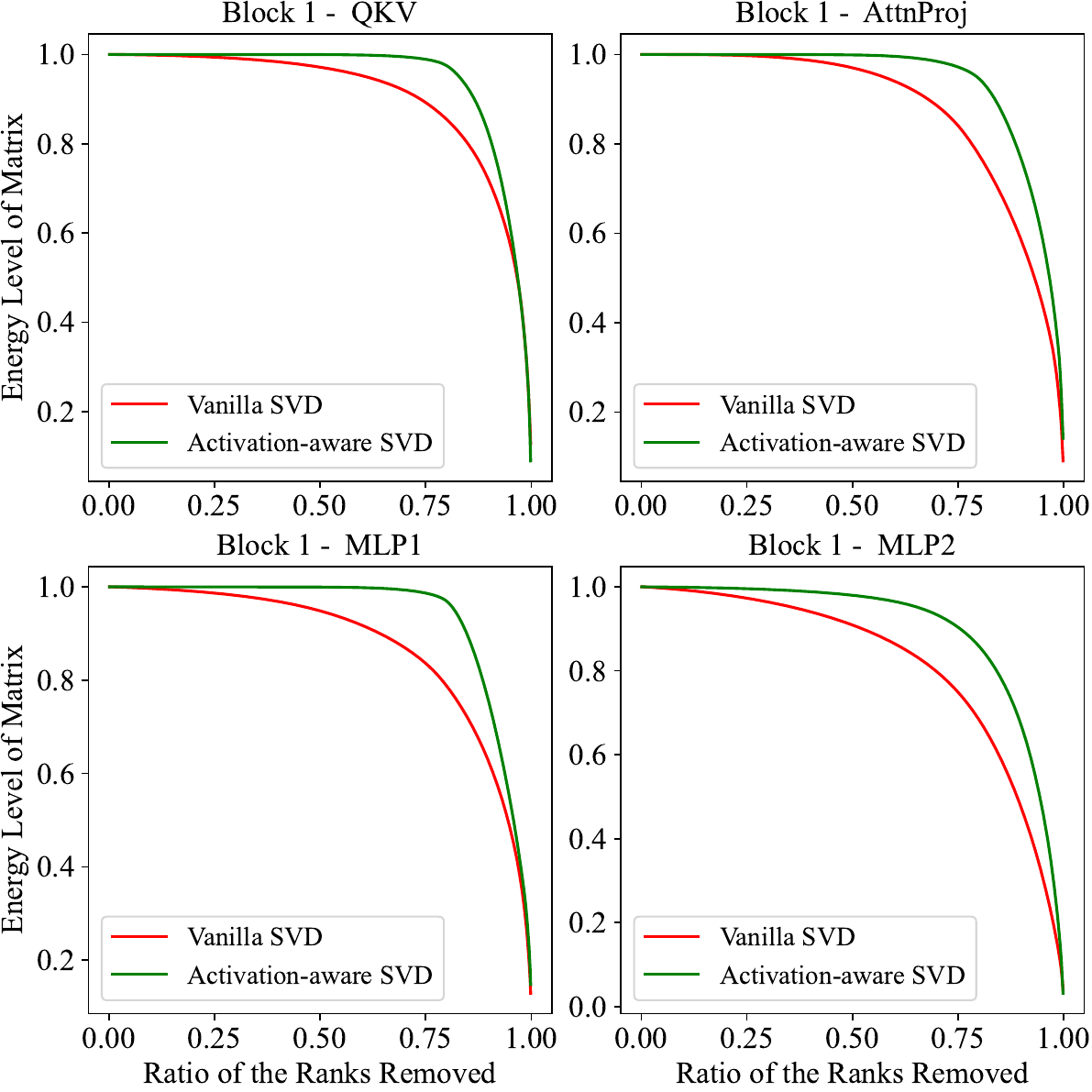}

    \caption{Impact of SVD-based rank reduction on energy level of different matrices in the first block of DeiT-B.}
    \label{fig:actsvd}

\end{figure}

The approach we have taken in applying SVD aligns with the findings of \cite{DBLP:conf/aaai/YuW23}, particularly regarding the layer's output being more amenable to low-rank approximation compared to its weights. The matrix energy, as defined above, serves as a critical metric in this context. In Fig.~\ref{fig:actsvd}, we analyze the energy loss in various layers of the DeiT-B model, comparing our activation-aware SVD approach to the direct application of SVD on weight matrices.
Results indicate that activation-aware SVD is more effective in preserving the matrix energy, especially for higher rank reduction factors. 

\subsection{Mixed-Rank Model Compression} \label{MRMC}
To achieve a specific model compression factor \(\alpha\), regardless of the type of low-rank approximation employed (i.e., whether it is activation-aware or conventional SVD), it is possible to assign the ranks non-uniformly across the layers. This approach allows for a more flexible and potentially more effective compression strategy.

\begin{figure}[tb]
    \centering
    
    \subfloat[Reducing the rank of a single block] {
        \includegraphics[width=0.48\columnwidth, valign=c]{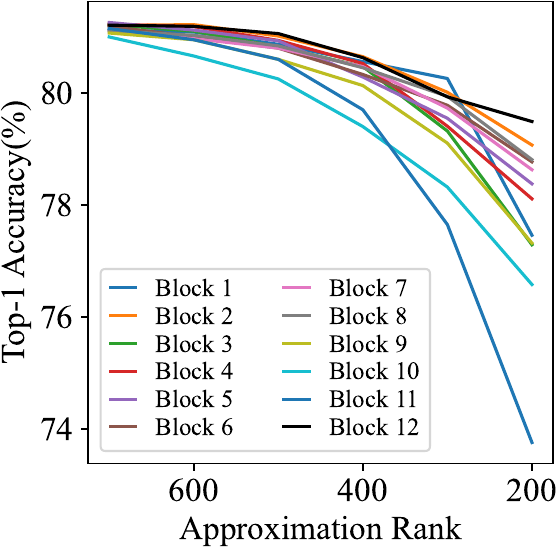}
        \label{fig:svd-sub-layer}
    }
    \subfloat[Reducing the rank of all similar layers] {
        \includegraphics[width=0.48\columnwidth, valign=c]{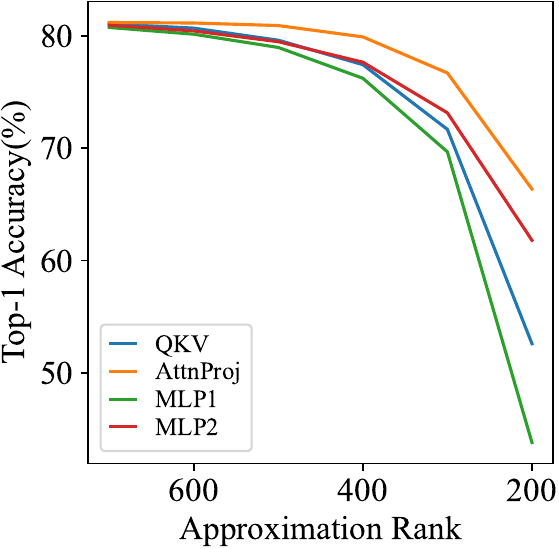}
        \label{fig:svd-all-sub-layers}
    }
    
    \caption{Impact of rank reduction on top-1 ImageNet accuracy}
    \label{fig:svd}
\end{figure}

In Fig.~\ref{fig:svd}, we present the relative decrease in energy of the weight matrices as the number of retained ranks in their spectrum is reduced.
This figure points to the necessity of assigning different ranks to various transformer blocks and even to individual layers within each block to preserve the model's performance effectively. This differentiation in rank allocation is crucial for maintaining a balance between model compression and performance retention. This phenomenon also aligns with the findings of the other domains of compression, including mixed-precision quantization, where layers of the model are differently sensitive to low bit-width quantization \cite{DBLP:conf/cvpr/WangLLLH19, azizi2023sensitivity, DBLP:conf/iccv/DongYGMK19}.  

Considering Problem \eqref{mainproblem}, our goal is to identify an optimal rank configuration \(\mathbold{r} \), which not only reduces the model's parameter count by a factor of \(\alpha\) but also minimizes the sum of energy losses observed across all layers of the model. As discussed in section \ref{act-aware lra}, we have already presented a technique that, for a given rank \(r_l\), calculates the optimal energy loss \(\varepsilon_l\) at the output of each layer \(l\). So, we can simplify the formulation of \eqref{mainproblem} as follows:
\begin{equation}
\begin{aligned}
\label{simpleproblem}
\mathrm{Err}(\mathcal{M}) =  \min_{\mathbold{r} } &\sum_{l}\varepsilon_l \\
\textrm{s.t.} \quad & \psi(\mathcal{M},\mathbold{r} ) \geq \alpha \\
\end{aligned}
\end{equation}

As we discussed earlier in section \ref{method}, this is still an NP-hard problem. To address this, we propose an efficient greedy solution based on local neighborhood search. Instead of a drastic, single-step reduction in ranks and parameters, we adopt a gradual optimization strategy. This method involves incremental steps, each of which slightly reduces the number of parameters in line with a pre-defined \textbf{compression rate scheduling policy}. For each step of parameter reduction, we select the layer whose (further) rank reduction yields the \textbf{minimum normalized energy loss} compared to its uncompressed counterpart, and reduce its rank to meet the target parameter count reduction. This iterative process avoids disproportionate normalized energy loss at any individual layer, thus greedily minimizing the sum of normalized energy loss across all layers (see \eqref{mainproblem}).

Algorithm~\ref{alg:example} details this process where function \textbf{ActivationAwareSVD()} is what we developed in section \ref{act-aware lra} for approximating \(\varepsilon_l\), and \(\tau\) is the rate scheduling function that sets the rate of rank reduction.
\begin{algorithm}[tb]
   
    \caption{Mixed-Rank Compression}
    \label{alg:example}
    \begin{algorithmic}[1]
   
        \Require Model $\mathcal{M}$, proxy dataset $\mathcal{D}$, Compression ratio $\alpha$
        \Ensure List of ranks $\mathbold{r}$
        \State Singular Values $\mathbold{S} = \mathrm{ActivationAwareSVD}(\mathcal{M}, \mathcal{D})$
        \State Initialize $p_\mathrm{tot} = p_\mathrm{cur} = \sum_l{n_l\times d_l}$ \Comment{ \# of param.}
        \State Initialize $\mathbold{r}^0 = \min(n_l, d_l) \; \forall l$ \Comment{ hidden dim. in ViT}
        \State $t = 1$
        
        \Repeat
            \State $p^t = \tau(\mathcal{M}, t, \alpha)$  \Comment{ \# of param. to remove at step $t$}
            \State $\mathbold{\ell}=[]$  \Comment{ vector for keeping energy losses}
            \State $\mathbold{m}^t=[]$  \Comment{ vector for keeping temporary ranks}   
            \State $\mathbold{r}^t=\mathbold{r}^{t-1}$
            \For{$l=1$ {\bfseries to} $L$}
                \State $\mathbold{m}^t[l] = \mathbold{r} ^{t-1}[l] -\cfrac{p^t}{n_l+d_l}$  \Comment{ \# of ranks to keep}
                \State $\mathbold{\ell}[l] = \mathrm{ComputeEnergyLoss}(\mathbold{S}[l], \mathbold{m}^t[l])$
            \EndFor
            \State $l^*=\argmin_l \mathbold{\ell}$  \Comment{ layer with minimum energy loss}
            \State $\mathbold{r} ^{t}[l^*] = \mathbold{m}^t[l^*]$
            \State $p_\mathrm{cur} = p_\mathrm{cur} - p^t$
            \State $t = t + 1$
        \Until{$p_\mathrm{cur} \leq \frac{p_\mathrm{tot}}{\alpha}$}
        \State \Return $\mathbold{r} ^{t}$
    \end{algorithmic}
\end{algorithm}
We adopt an exponential of the form below:
\begin{equation}
    \tau(\mathcal{M},iter, \alpha) = N_{target} + (N_0- N_{target})\exp{(-\frac{iter}{\gamma})}
\end{equation}
in which \(N_0\) is the initial number of parameters, \(N_{target}\) is our desired number of parameters, and \(\gamma\) is the decaying rate of the schedule. The exponential nature of the schedule encourages significant parameter reduction in the initial stages, primarily targeting layers that are less sensitive to rank reduction. As the process progresses, the required rate of parameter reduction is decreased like an annealing schedule.
Consequently, this approach minimizes the impact on layers that are more sensitive to rank reduction.
The function \textbf{ComputeEnergyLoss()}, which is called for each layer \(l\) and in each iteration \(t\) of algorithm, computes the ratio of the lost energy for layer \(l\) and its original total energy if it only retains \(\mathbold{m}^t[l]\) of its singular values. 

A key aspect of Algorithm~\ref{alg:example} is its efficiency in handling the Singular Value Decomposition (SVD) of each layer. Notably, the SVD for each layer needs to be computed only once. After this initial computation, we store the singular values obtained from the decomposition. Subsequently, all optimization processes leverage these stored singular values. This approach significantly reduces computational overhead, as it eliminates the need for repeated SVD computations for each layer during the optimization steps.

\subsection{Layer-Wise Error Compensation}
\label{gbopt}
Using the combination of the techniques developed in sections \ref{act-aware lra} and \ref{MRMC} we able to come up with the low-rank approximation matrices \(\mathbold{U}\) and \(\mathbold{V}\) for each layer, with the ranks specific to each layer.  However, applying SVD and removing some of the singular vectors would unavoidably introduce some energy loss. In mixed-rank compression  we managed to create a balanced distribution of energy losses across all layers. In this section, we present a novel and efficient method to compensate for the energy loss experienced by all layers effectively. Revisiting \eqref{layerwise}, we are looking for an auxiliary matrix \(\mathbold{Z}\) that is efficient in terms of the number of parameters and can compensate for the residual error introduced by activation-aware SVD.

We investigate four distinct configurations for the \(\mathbold{Z}\) matrix: (1, 2) Sparse matrix (both structured and unstructured) in a manner similar to \cite{DBLP:conf/icml/LiYZLHCZ23}, where sparse matrices were utilized for capturing the uncorrelated components of the approximation's residual; (3) Diagonal matrix; (4) Low-rank matrix, with a rank substantially smaller than that of \(\mathbold{U}\) and \(\mathbold{V}\). Table~\ref{tab:comp_app} compares these strategies in terms of the parameter count overhead and normalized remaining residual error after the layer-wise error compensation.
\begin{table}[tb]
    \centering
    \caption{Comparison of different approaches for residual error compensation via \(\mathbold{Z}\).}
    \label{tab:comp_app}
    
    \begin{flushleft} \scriptsize
        \(\uparrow\) and \(\downarrow\) indicates relatively high and low values, respectively.
    \end{flushleft}
    
    \resizebox{0.8\columnwidth}{!}{
    \begin{tabular}{lcc}
        \toprule
        \multirow{2}{*}{Approach} & Parameter Count &       Average       \\
        {}                        &    Overhead     &   Residual Error    \\ \midrule
        Sparse (unstructured)      &  \(\uparrow\)   & 0.11 \(\downarrow\) \\
        Sparse (structured)        & \(\downarrow\)  &  0.23 \(\uparrow\)  \\
        Diagonal                  & \(\downarrow\)  &  0.27 \(\uparrow\)  \\
        \textbf{Low-Rank}         & \(\downarrow\)  & 0.13 \(\downarrow\) \\ \bottomrule
    \end{tabular}
    }
    
\end{table}
Low-rank matrix is a compelling option regarding both metrics, thus we introduce \(\mathbold{G}\mathbold{Y}^T\) as the matrix for capturing the residual error, where \(\mathbold{G} \in \mathbb{R}^{n\times q}\) and \(\mathbold{Y} \in \mathbb{R}^{d \times q}\) have a rank much smaller than that of \(\mathbold{U}\) and \(\mathbold{V}\) (\(q \ll r_l  \; \forall l\)).

The introduction of \(\mathbold{G}\) and \(\mathbold{Y}\) serves as a key innovation for addressing the energy loss inherent in the SVD process. By starting with these matrices as zero and progressively updating them, we are able to fine-tune the approximation in a way that specifically targets the reconstruction errors and energy deficiencies resulting from the initial SVD. Firstly, let's revisit \eqref{layerwise} with a slight modification. In this iteration, we maintain \(\mathbold{U}\) and \(\mathbold{V}\) as fixed, based on the values obtained from the activation-aware low-rank approximation (section \ref{act-aware lra}). Additionally, we substitute the expectation term in the equation with a summation over the samples in the proxy dataset. This adjustment aligns with our earlier discussion about the computational challenges of directly computing the expectation and the practical solution of using a proxy dataset \(\mathcal{D}\) for approximation:

\begin{equation}
    \label{layerwise1}
    \footnotesize
    \varepsilon_{l}=\min_{\mathbold{G}_l,\mathbold{Y}_l}\frac{1}{|\mathcal{D}|}\sum_{i}\left[\frac{\parallel \mathbold{X}_l^i(\mathbold{G}_l\mathbold{Y}_l^T) - \mathbold{X}_l^i(\mathbold{W}_l-\mathbold{U}_l\mathbold{V}_l^T) \parallel_F^2}{\big(\parallel \mathbold{X}_l^i\mathbold{W}_l \parallel_F^2 \big)}\right] 
\end{equation}

To simplify the implementation process, we choose to fix the rank \( q \) for each layer statically. This approach entails setting the total number of parameters in \( \mathbold{G}_l \) and \( \mathbold{Y}_l \) to a pre-defined percentage of the parameters in \( \mathbold{W}_l \). We opt for this percentage to be 5\%.

With the dimensions of these low-rank matrices \(\mathbold{G}_l\) and \(\mathbold{Y}_l\) fixed, the problem essentially transforms into a regression task. For each layer \(l\), we define \(\mathbold{A}_l^i=\frac{\mathbold{X}_l^i}{\big(\parallel \mathbold{X}_l^i\mathbold{W}_l \parallel_F^2 \big)}\), \(\mathbold{B}_l^i = \frac{\mathbold{X}_l^i(\mathbold{W}_l-\mathbold{U}_l\mathbold{V}_l^T)}{\big(\parallel \mathbold{X}_l^i\mathbold{W}_l \parallel_F^2 \big)}\), and the loss function \(\mathcal{L}_l =\frac{1}{|\mathcal{D}|}\sum_{i}\| \mathbold{A}_l^i(\mathbold{G}_l\mathbold{Y}_l^T) - \mathbold{B}_l^i \|_F^2\). This setting forms the basis of our optimization problem. A crucial observation is that this optimization process can be conducted for each layer of the model independently. As a result, the optimization tasks for different layers can be executed \textbf{in parallel}. This parallelization greatly enhances the efficiency of the optimization process, allowing for simultaneous adjustments across multiple layers. Although there is no closed-form solution, this problem can be solved using gradient descent since the gradient of the loss function with respect to \(\mathbold{G}\) and \(\mathbold{Y}\) can be easily computed as below:

\begin{equation}
    \footnotesize
    \label{uderivative}
    \frac{\partial \mathcal{L}}{\partial \mathbold{G}} = \frac{1}{|\mathcal{D}|}\sum_{i}2{\mathbold{A}^i}^T(\mathbold{A}^i\mathbold{G}\mathbold{Y}^T-\mathbold{B}^i)\mathbold{Y} 
\end{equation}
\begin{equation}
    \footnotesize
    \label{vderivative}
    \frac{\partial \mathcal{L}}{\partial \mathbold{Y}} = \frac{1}{|\mathcal{D}|}\sum_{i}2(\mathbold{A}^i\mathbold{G}\mathbold{Y}^T-\mathbold{B}^i)^T\mathbold{A}\mathbold{G}
\end{equation}
Intuitively, the \(\mathbold{X}\mathbold{U}\mathbold{V}^T\) component is designed to capture the main components and a substantial portion of the energy of \(\mathbold{X}\mathbold{W}\) via activation-aware SVD. Consequently, \(\mathbold{X}\mathbold{G}\mathbold{Y}^T\) endeavors to minimize the residual error left by the SVD, utilizing the relatively small matrices \(\mathbold{G}\) and \(\mathbold{Y}\). This dual approach aims to ensure a thorough and efficient approximation of the original weight matrices.

\begin{figure}[tb]
    \centering
    \includegraphics[width=0.65\columnwidth]{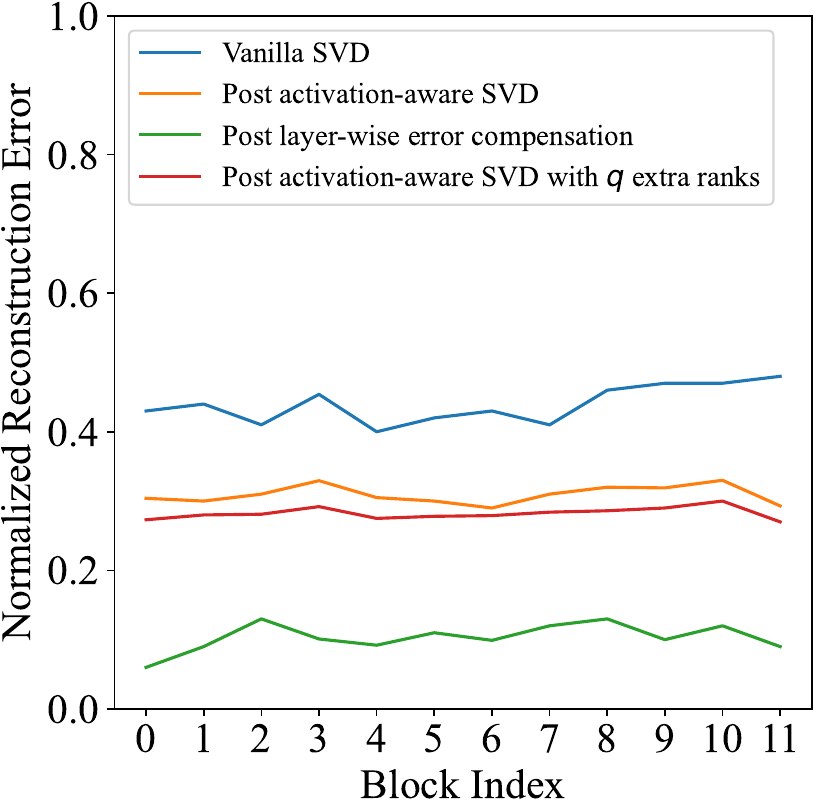}

    \caption{Normalized Frobenius norm of the error at the output of the AttnProj layer in DeiT-B}
    \label{fig:errors}

\end{figure}

To demonstrate the effectiveness of this addition, we analyze the Frobenius norm error of the attention projection module in DeiT-B model. This analysis is conducted post-activation-aware SVD, following the gradient-based optimization on an unseen (validation) proxy dataset. The results, illustrated in Fig. \ref{fig:errors}, highlight the impact of our approach on these layers, showcasing the reduction in reconstruction error achieved through our optimization process.
The activation-aware SVD achieves an almost uniform distribution of error across the layers. This outcome aligns perfectly with the primary objective of our problem; on top of that, the gradient-based optimization  reduces the layer-wise error as much as possible. If we omit the gradient-based error compensation and instead allocate the parameter budget of \(\mathbold{G}\) and \(\mathbold{Y}\) to \(\mathbold{U}\) and \(\mathbold{V}\), the error reduction would be a lot less than the residual error reduction through \(\mathbold{G}\) and \(\mathbold{Y}\). This scenario highlights the critical role of error compensation, as evidenced by the comparative results.
An important aspect of this optimization is that \(\mathbold{U}_l\mathbold{V}_l^T\) remains constant while optimizing \(\mathbold{G}_l\mathbold{Y}_l^T\). Given that the dimensions of \(\mathbold{G}\) and \(\mathbold{Y}\) are significantly smaller than those of \(\mathbold{U}\) and \(\mathbold{V}\), there is a minimal risk of overfitting to the calibration dataset, thus providing more generalizability and robustness.

\section{Results and Discussions}
In this section, we thoroughly evaluate our compression framework. 
\subsection{Experimental Setup}
In our experimental setup, we have carefully selected specific hyperparameters to optimize the performance of our model. Here is an overview of the key settings:

1. Proxy Dataset Creation: For both activation-aware SVD and gradient-based optimization, we construct the proxy dataset using 1024 samples randomly selected from the dataset. This sample size is chosen to provide a representative subset of the original data.

2. Rank Setting for \(\mathbold{G}\) and \(\mathbold{Y}\): The rank \(q\) for the matrices \(\mathbold{G}\) and \(\mathbold{Y}\) is determined so that these matrices together account for 5\% of the parameters in each layer.

3. Mixed-Rank Gradual Compression: In this approach, we set the decay rate \(\gamma\) at 80 and the iterative algorithm for this compression is run for 500 iterations.

4. Layer-wise Error Compensation Settings (Section \ref{gbopt}): During the gradient-based optimization phase, we set the learning rate for updating \(\mathbold{G}\) and \(\mathbold{Y}\) to \(10^{-3}\). This process involves running Mini-Batch gradient descent for 2000 iterations with a batch size of 64. Since this optimization is done in a layer-wise manner, no backpropagation is involved, and layers can be processed in parallel; thus, it is very fast.

5. As the final step, we fine-tune the uncompressed parameters of the model, including the \textbf{LayeNorm} parameters, \textbf{head}, \textbf{biases}, and the \textbf{patch embedding} module on the standard ImageNet dataset. This \textbf{partial fine-tuning} is done for 20 epochs with a learning rate of  \(10^{-4}\) and a cosine scheduling. Since almost \(~2\%\) of the network parameters are being fine-tuned and the weight matrices are freezed (they are already optimized using our flow), this step is very fast and efficient.

6. Library and Hardware: We utilize pre-trained models from the timm library \cite{rw2019timm} and implement our model optimization using PyTorch \cite{DBLP:conf/nips/PaszkeGMLBCKLGA19}. All experiments are conducted on NVIDIA A6000 GPUs.

\subsection{ImageNet Classification}
We assess our framework's effectiveness in compressing various Vision Transformer architectures, including ViT \cite{DBLP:conf/iclr/DosovitskiyB0WZ21}, DeiT \cite{DBLP:conf/icml/TouvronCDMSJ21}, and Swin Transformer \cite{DBLP:conf/iccv/LiuL00W0LG21}, on the ImageNet dataset. The detailed results are presented in table \ref{tab:results}.
The table includes Params, indicating both the absolute number of parameters and the percentage of reduction relative to the baseline model. We evaluate the models at three different levels of parameter count reduction: -40\%, -50\%, and -60\%, and correspondingly report the Top-1 validation accuracy.

\begin{table}[tb]
\centering
\caption{Comparison of the parameter count and top-1 ImageNet accuracy for different compression methods and ViT architectures.}
\begin{flushleft} \scriptsize
FT stands for fine-tuning.\\
The results of other methods are directly sourced from their respective papers (where available): AAFM \cite{DBLP:conf/aaai/YuW23}, WDPruning \cite{DBLP:conf/aaai/0004HWCCC22}, SPViT \cite{DBLP:conf/eccv/KongDMMNSSYRTQW22}, \(\mathrm{S^2ViTE}\) \cite{DBLP:conf/nips/ChenCGYZW21}, and VTP \cite{zhu2021vision};
\end{flushleft}
\label{tab:results}
\resizebox{\columnwidth}{!}{
\begin{tabular}{c l c c}
    \toprule
                 \multirow{2}{*}{\textbf{Architecture}}               & \multirow{2}{*}{\textbf{Method}} & \textbf{\# Params}     & \textbf{Top-1}         \\
                                   {}                                 & {}                               & \textbf{(Millions)}    & \textbf{Accuracy (\%)} \\ \midrule[\heavyrulewidth]
                    \multirow{9}{*}{\textbf{DeiT-B}}                  & Baseline                         & {86.6}                 & {81.80}                \\
                           \cmidrule{2-4}
 {}                         & {AAFM}                           & {51.9(-40\%)}          & {81.28}                \\
                                   {}                                 & {WDPruning}                      & {60.6(-30\%)}          & {81.10}                \\
                                   {}                                 & {\(\mathrm{S^2ViTE}\)}           & {56.8(-35\%)}          & {82.20}                \\
                                   {}                                 & {SPViT}                          & {62.3(-28\%)}          & {81.60}                \\
                                   {}                                 & {Vanilla SVD + FT}               & {52.0(-40\%)}          & {77.30}                \\
                                   {}                                 & {\textbf{Ours}}                  & {\textbf{52.0(-40\%)}} & {\textbf{81.80}}       \\
                                   {}                                 & {\textbf{Ours}}                  & {\textbf{43.3(-50\%)}} & {\textbf{81.35}}       \\
                                   {}                                 & {\textbf{Ours}}                  & {\textbf{34.6(-60\%)}} & {\textbf{81.10}}       \\
    \cmidrule[\heavyrulewidth]{1-4}
 \multirow{8}{*}{\textbf{DeiT-S}} & Baseline                         & {22.1}                 & {79.80}                 \\
                           \cmidrule{2-4}

 {}                        & {WDPruning}                      & {15.0(-32\%)}          & {78.60}                \\
                                   {}                                 & {\(\mathrm{S^2ViTE}\)}           & {14.6(-34\%)}          & {79.20}                \\
                                   {}                                 & {SPViT}                          & {16.4(-26\%)}          & {78.30}                \\
                                   {}                                 & {Vanilla SVD + FT}               & {13.2(-40\%)}          & {75.20}                \\
                                   {}                                 & {\textbf{Ours}}                  & {\textbf{13.2(-40\%)}} & {\textbf{79.60}}       \\
                                   {}                                 & {\textbf{Ours}}                  & {\textbf{11.1(-50\%)}} & {\textbf{79.34}}       \\
                                   {}                                 & {\textbf{Ours}}                  & {\textbf{8.9(-60\%)}}  & {\textbf{78.60}}        \\
    \cmidrule[\heavyrulewidth]{1-4}
 \multirow{5}{*}{\textbf{ViT-B}}  & Baseline                         & {86.5}                 & {84.55}                 \\
                            \cmidrule{2-4}
{}                         & {VTP}                            & {48.0(-45\%)}          & {80.70}                \\
                                   {}                                 & {Vanilla SVD + FT}               & {42.3(-50\%)}          & {78.50}                \\
                                   {}                                 & {\textbf{Ours}}                  & {\textbf{52.9(-40\%)}} & {\textbf{84.20}}       \\
                                   {}                                 & {\textbf{Ours}}                  & {\textbf{42.3(-50\%)}} & {\textbf{83.70}}       \\
                                   {}                                 & {\textbf{Ours}}                  & {\textbf{34.6(-60\%)}} & {\textbf{83.10}}       \\ 
    \cmidrule[\heavyrulewidth]{1-4}
 \multirow{7}{*}{\textbf{Swin-B}} & Baseline                         & {88.1}                 & {85.45}                 \\
                           \cmidrule{2-4}
 {}                         & {AAFM}                           & {60.2(-33\%)}          & {82.68}                \\
                                   {}                                 & {SPViT}                          & {68.0(-24\%)}          & {83.20}                \\
                                   {}                                 & {Vanilla SVD + FT}               & {52.9(-40\%)}          & {79.10}                \\
                                   {}                                 & {\textbf{Ours}}                  & {\textbf{52.9(-40\%)}} & {\textbf{83.90}}       \\
                                   {}                                 & {\textbf{Ours}}                  & {\textbf{44.1(-50\%)}} & {\textbf{83.65}}       \\ 
                                   {}                                 & {\textbf{Ours}}                  & {\textbf{35.3(-60\%)}} & {\textbf{83.14}}       \\ \bottomrule
\end{tabular}}
\end{table}

Our results demonstrate that we can achieve a 50\% reduction in the parameter count of DeiT-B without any loss in accuracy. Furthermore, we accomplish a 60\% reduction in parameters with less than a 1\% drop in accuracy, showcasing the efficacy of our compression framework across different levels of parameter reduction. It can be seen that our model outperforms the previous compression approaches in accuracy, compression, or both. 

To demonstrate our method's capabilities, we undertake extreme compression cases: compressing ViT-L to match DeiT-B's size, DeiT-B to DeiT-S's size, and DeiT-S to DeiT-T's size. This tests our method's efficacy in significantly reducing model sizes while maintaining performance. The outcomes of these compression experiments are detailed in table \ref{tab:moreresults}.
The data shows that our compressed models, despite having a similar number of parameters, outperform the pre-trained models they are compared against, highlighting the effectiveness of our compression method.

\begin{table}[tb]
    \centering
    \caption{Comparison of the top-1 ImageNet accuracy on models with approximately the same size.}
    \label{tab:moreresults}

    \resizebox{0.9\columnwidth}{!}{
    \begin{tabular}{l c c}
        
        \toprule
        \multirow{2}{*}{\textbf{Architecture}} & \multirow{2}{*}\textbf{\# Params}         & \textbf{Top-1}  \\
        
        {}                                      &
         \textbf{(Millions)}     & \textbf{Accuracy (\%)}\\
        
        \midrule
        ViT-B                        & {86.6}            & {81.8}  \\
        \textbf{Compressed ViT-L}    & {\textbf{86.0 (-72\%)}}    & {\textbf{83.1}}     \\
        \midrule
        
        DeiT-S                      & {22.1}    & {79.8}  \\
        \textbf{Compressed DeiT-B}  & {\textbf{21.7 (-75\%)}}    & {\textbf{80.3}}     \\
        \midrule
        
        DeiT-T                      & {5.0}    & {72.2}  \\
        \textbf{Compressed DeiT-S}  & {\textbf{5.0 (-77\%)}}    & {\textbf{74.0}}     \\

        \bottomrule
    \end{tabular}
    }
\end{table}

\begin{table}[ht]
\centering
\caption{Compatibility of the presented compression approach with weight quantization.}
\label{tab:compatibility}
\resizebox{\columnwidth}{!}{
\begin{tabular}{l c c}
    \toprule
    \multirow{2}{*}{\textbf{Method}}                              & \textbf{Model}      & \textbf{Top-1}         \\
    {}                                                            & \textbf{Size (MiB)} & \textbf{Accuracy (\%)} \\
    \cmidrule[\heavyrulewidth]{1-3}
    DeiT-B baseline (FP16)                                        & {165.2}             & {81.80}                \\
    DeiT-B 50\% compressed (FP16)                                 & {82.6}                & {81.35}                \\
    \textbf{DeiT-B 50\% compressed + 8-bit PTQ}                            & \textbf{41.7}                & \textbf{81.15}                \\
    DeiT-B baseline + 4-bit PTQ                                        & {42.2}             & {80.72}                \\
    \bottomrule
\end{tabular}}
\end{table}

\subsection{Compatibility with Weight Quantization}

To illustrate the compatibility of the presented approach with weight quantization, we apply post-training quantization (PTQ) to weight matrices \(\mathbold{U}\), \(\mathbold{V}\), \(\mathbold{G}\), and \(\mathbold{Y}\). For this purpose, we employ 8-bit, channel-wise, round-to-nearest quantization, targeting only the weights (the details of the quantization function applied can be found in appendix). We then compare this mixed compression strategy with applying 4-bit PTQ to the baseline model, which yields about the same level of compression.

As summarized in table \ref{tab:compatibility}, the 8-bit version of our compressed model surpasses the 4-bit uncompressed model in accuracy by 0.43\%. This result indicates the compatibility of quantization with our low-rank approximation and its superiority compared with quantization-only compression.


\section{Conclusions and Future Work}
In this work, we proposed a novel methodology for compressing ViTs. We adopted activation-aware SVD to approximate the outputs of the layers within the model while maintaining the principal energy components of the matrices. This approximation was refined by developing a greedy strategy for assigning various ranks to different layers. In the end, we also proposed layer-wise error compensation for reducing the error introduced by SVD as much as possible. Overall, our methodology significantly reduces the parameter count of ViTs, facilitating their efficient deployment in inference engines.

While our current work has been focused on the compression of ViTs, the presented approach exhibits promising characteristics that suggest its potential applicability to other transformer-based architectures like large language models.
Investigating and adapting our methodology to other transformer variants is an exciting direction for future work.

\clearpage

\nocite{langley00}

\bibliography{example_paper}
\bibliographystyle{icml2024}

\newpage
\appendix
\onecolumn

\section{Impact of Low-Rank Approximation on Compute Efficiency}
\label{flops}

Given a matrix multiplication \(\mathbold{O} = \mathbold{X}\mathbold{W}\), where \(\mathbold{X}\) and \(W\) are \(n \times k\) and \(k \times m\) matrices respectively, we approximate the output as \(\mathbold{O} \approx \mathbold{X}(\mathbold{U}\mathbold{V}^T + \mathbold{G}\mathbold{Y}^T)\). Here, \(\mathbold{U}\), \(\mathbold{V}\), \(\mathbold{G}\), and \(\mathbold{Y}\) are \(k \times r\), \(m \times r\), \(k \times q\), and \(m \times q\) matrices, respectively. The original matrix multiplication requires \(nkm\) individual multiplication operations.

If we compute \(\mathbold{U}\mathbold{V}^T\) and \(\mathbold{G}\mathbold{Y}^T\) directly, add them, and then multiply by \(\mathbold{X}\), the total number of multiplications is:
\begin{equation}
    krm \; (\text{for} \; \mathbold{U}\mathbold{V}^T) + kqm \; (\text{for} \; \mathbold{G}\mathbold{Y}^T) + nkm \; (\text{when multiplying by} \; \mathbold{X})
\end{equation}
This increases the multiplication count. However, a more efficient approach is to first multiply \(\mathbold{X}\) by \(\mathbold{U}\) and then the result by \(\mathbold{V}^T\), and similarly for \(\mathbold{G}\) and \(\mathbold{Y}\). Compute \(\mathbold{X}\mathbold{G}\) first, then multiply it by \(\mathbold{Y}^T\). The total multiplications become:
\begin{equation}
    nkr \; (\text{for} \; \mathbold{X}\mathbold{U}) + nrm \; (\text{for} \; \mathbold{X}\mathbold{U}\mathbold{V}^T) + nkq \; (\text{for} \; \mathbold{X}\mathbold{G}) + nqm \; (\text{for} \; \mathbold{X}\mathbold{G}\mathbold{Y}^T)
\end{equation}
The ratio of these multiplications to the original is:
\begin{equation}
    \frac{nkr + nrm + nkq + nqm}{nkm} = \frac{nk(r+q) + mn(r+q)}{nkm}
\end{equation}
Since \(r+q\) is smaller than the original matrix dimensions \(k\) and \(m\), this ratio is less than 1. Thus, our method, while not primarily aimed at computational efficiency, inadvertently achieves a reduction in multiplication count, leading to a speedup.

\section{Quantization}
\label{quant}
In our experiments, post-training quantization (PTQ) was implemented to demonstrate how our weight compression technique can be effectively combined with other methods. The specific quantization function used is the basic round-to-nearest linear quantization, defined as follows:
\begin{equation}
    \mathbold{U}_q = \mathrm{clamp}(\lfloor\frac{\mathbold{U}}{s}+z\rceil, 0, 2^N-1), \quad \hat{\mathbold{U}} =  s \times (\mathbold{U}_q-z)
\end{equation}
Here, \(s = \frac{\max (\mathbold{U})}{2^N-1}\) and \(z = -\frac{\min (\mathbold{U})}{2^N-1}\), where \(N\) is the number of bit-widths used. The operation \(\lfloor \rceil\) represents the rounding process. \(\mathbold{U}_q\) is the quantized version, and \(\hat{\mathbold{U}}\) is the de-quantized version of the original matrix \(\mathbold{U}\). 

\end{document}